# Probabilistic Acceptance

Henry E. Kyburg, Jr.
Computer Science and Philosophy
University of Rochester, Rochester, NY 14627, USA
kyburg@cs.rochester.edu *

## Abstract

The idea of fully accepting statements when the evidence has rendered them probable enough faces a number of difficulties. We leave the interpretation of probability largely open, but attempt to suggest a contextual approach to full belief. We show that the difficulties of probabilistic acceptance are not as severe as they are sometimes painted, and that though there are oddities associated with probabilistic acceptance they are in some instances less awkward than the difficulties associated with other nonmonotonic formalisms. We show that the structure at which we arrive provides a natural home for statistical inference.

## 1 Introduction.

You and I often jump to conclusions that are not strictly (deductively) entailed by the evidence and background knowledge we have available. In doing so, we are not always acting irrationally. An alternative would be to assign to each proposition the degree of belief less than unity that is appropriate, in the light of the evidence, but life is too short to calculate these degrees of belief, even if they exist. Many writers, therefore, have been led to consider nonmonotonic inference: inference that goes beyond deduction, but suffers the drawback of occasionally leading to falsehood from true premises. For present purposes we skip the important debate between "probabilists" and "logicists" and simply observe that many people take the human propensity to jump to conclusions to be a potentially valuable ingredient of artificial cognitive systems. It is this conviction that has driven the development of such nonmonotonic systems as autoepis-

* Research for this work was supported by the National Science Foundation, grant IRI-9411267

temic logic [Moore, 1985], default logic [Reiter, 1980], theorist [Poole, 1991], defeasible logic [Pollock, 1987; Loui, 100 106], circumscription [McCarthy, 1980], and many others.

One natural gloss on nonmonotonic or uncertain inference is to say that we accept what is probable. It is well known that this can't work. Our main purpose here is to show that this particular bit of folk wisdom — that acceptance on the basis of probability cannot be taken as the foundation of uncertain inference — is mistaken.

## 2 Preliminaries

To make this thesis more precise requires getting clearer about what we mean by *acceptance*, what we mean by *probability*, and what we can reasonably demand of a system of uncertain inference.

We can be quite general about probability. We need only require that it be a function whose domain includes closed sentences of our language and whose range, whether it be real numbers, intervals of reals, sets of reals, or fuzzy sets of reals, be such that the idea of a real-valued threshold makes sense: $P(S) \geq t$. If $P(S)$ is an interval or a set of reals, then $P(S) \geq t$ means that the *lower bound* of $P(S)$ is greater than $t$. In particular, in interpreting probability, we can leave open the question of whether all probabilities are "ultimately" based on objective statistics (as we believe) or whether some or all probabilities are essentially subjective in character. (Note that we cannot *identify* probability with frequency. To do so would require that we take the probability of heads on the next toss to be 0 or 1, since there is only one next toss and it either yields heads or it does not; no other frequencies are admissible.)

The idea of acceptance requires somewhat more detailed consideration. Clearly we intend it to be tentative, or nonmonotonic. On the other hand, acceptance must be distinguished from merely having a certain degree of belief. One suggestion [Braithwaite, 1946] is that to *accept* a proposition is to be prepared to *act* on the basis of its truth, as opposed to being prepared to *bet* on its truth. This doesn't seem quite right, since given any proposition, it is usually possible to conjure up bizarre circumstances under which one would not be prepared to act as if it were true [Levi and Morgenbesser, 1964].

A more reasonable idea is to construe acceptance as *somewhat* context relative. That is, when we talk of acceptance we have in mind some range of circumstances (e.g., planning a trip by public transportation; deciding which of a certain limited set of acts to perform on the basis of one or another possible body of evidence; etc.) *within* which an accepted proposition is to be regarded as true. Another way to put this is to say that within this range of circumstances, we do not take an accepted proposition as a suitable matter for a bet [Kyburg, 1988].

For example, suppose that in the decisions you face in a certain class of circumstances the ratio of costs to benefits always lies between 1 : 3 and 3 : 1. In this class of circumstances there is no difference between a probability of 0.75 and a probability of 1.0, and no difference between a probability of 0.25 and a probability of 0.0. Even holding the class of circumstances constant, however, acceptance is nonmonotonic. Given evidence $E$, the proposition $S$ may be acceptable relative to the class $C$ of circumstances. We act as if $S$ is true. There are no odds we can encounter in $C$ that would lead us to bet against $S$. But when $E$ is enlarged by the addition of new information $F$, to yield total evidence $E \cup F$, then $S$ may no longer be accepted, even in $C$: we will no longer act as if $S$ is true in $C$; we may find circumstances in $C$ in which we would bet against $S$, etc.

This view of acceptance fits in reasonably well with the approach of nonmonotonic logic. When you know of Tweety only that she is a bird, you act as if that were true: you put a top on the cage, you don't — in ordinary contexts — bet that Tweety can't fly, etc. "Ordinary contexts": if some shifty-eyed character sidles up to you and offers to bet two to one that Tweety can't fly, you take that as relevant evidence that there is something going on that you don't know about. Similarly, if you don't know that you have a brother, you go ahead and act as if you don't, and you don't entertain bets about the matter. But it is not hard to imagine circumstances that would lead you to assign a degree of belief to that proposition rather than simply accepting it.

The upshot is that if the uncertainty is low enough, it is reasonable to act "on the basis of practical certainty" and to avoid the calculation of expected utility. On the other hand, if our uncertainty is not negligible, our action should be based on expected utility, and the "probabilities" that give rise to the expectation should be based on approximate frequencies of which we are "practically certain." In either case, there is a role for practical certainty in action.

## 3 Difficulties with Probabilistic Acceptance.

As natural as high probability is as a ground for tentative acceptance, probabilistic acceptance has received a lot of bad press. It has generally been dismissed as a ground of acceptance in the nonmonotonic world (except when "high" is taken to mean *arbitrarily* close to 1.0 [Adams, 1986; Geffner and Pearl, 1990].) This has been so for a number of reasons.

### 3.1  The Lack of Statistics.

Ever since the influential paper by Hayes and McCarthy [McCarthy and Hayes, 1969], it has been claimed that there are many natural instances of nonmonotonic inference that cannot be a matter of probability, "since the required statistical data are not available to the agent." Others have voiced similar arguments.

These arguments fail to hold water for two quite different reasons: some people hold that not all (or even no) probabilities *need* be based on statistical knowledge; and the problem of choosing the right reference class, when statistics *are* available is not as simple as these arguments suppose.

#### 3.1.1  Subjective Probability.

Recall that we left the interpretation of probability quite open — in particular we left open the possibility that not all probabilities need be based on statistical evidence. This means that, if we interpret probabilities as subjective, we can simply say that whenever, intuitively, the agent is entitled to infer $S$ from total evidence $T$, we are free to claim that the inference is warranted exactly because the agent is entitled to take the conditional probability of $S$ given $T$ to be high.

A number of writers [Pearl, 1992; Adams, 1986; Geffner and Pearl, 1990] have followed this line, but usually with the constraint that to justify acceptance, the conditional probability of $S$ given $T$ must be arbitrarily close to 1.0.

Few who adopt the currency of subjective probability are willing to squander it on acceptance, however. If belief comes in degrees, then perhaps we can explain all our rational decisions and actions in terms of degrees of belief that are less than the 1.0 that would characterize statements whose truth we have accepted. We need merely single out a class of statements to which we can assign probabilities of 1.0 on the basis of "observation" and take the correct epistemic attitude toward any statement to be its conditional probability, where the condition is our total observational knowledge. There are difficulties with this position (not the least of which is the problematic character of the term "observation") but for present purposes we leave this dispute to one side and assume that we need to make sense of acceptance for statements with probabilities less than 1.0.

### 3.1.2 Statistical Knowledge.

Generally, however, the force of the argument that many of the conclusions that we want to accept non-monotonically cannot be based on probability, depends on the implicit assumption that probabilities are to be based on statistical knowledge. From our point of view, this is a pretty plausible assumption. The objection is nevertheless off base, because there are so many sources of statistical knowledge, and there is more than one way in which a probability can be "based on" statistical knowledge.

1. We may have gathered the statistics of a large sample and inferred that they are characteristic of a population, of which the instance at hand is a member.

2. Some other dependable person may have gathered the data and reported it to us.

3. Some other dependable person may have inferred, from data gathered by yet other people, that a certain statistical generalization holds, and that person may have reported the statistical generalization to us.

4. The statistical generalization may be derived from other generalizations that in turn we obtain from reliable informants.

Note that in the last case particularly the generalization need not — and maybe cannot — be construed as representing a frequency in our world, much less as a direct generalization from an observable frequency in our world. It may represent a propensity in a possible world, or over a collection of possible worlds.

Furthermore, we should take a closer look at the sentences whose probabilities concern us. They may be of a form that leads quite directly to a statistic, such as the probability of "the next toss of *this* coin will land heads," where the coin is otherwise unspecified. We could imagine having a large store of data concerning *this* coin. But the sentences at issue may concern objects that are specified more precisely, such as "the next toss of this freshly minted, never-been-tossed, immediately-after-to-be-destroyed coin will land heads on its one and only toss." By its very characterization we cannot have a body of statistical data representing tosses of *that* coin.

Obviously, there are many ways of tying the next toss of that special coin to the reported and experimental history of coin tosses in general. One way is simply to point out that, epistemically, the toss described is, like the first toss, *merely* the toss of a coin [Kyburg, 1961]. Another approach would be to infer from the general statistical character of coin tosses, that in a possible world in which *this* coin *were* often tossed (as opposed to this world, in which it is tossed but once), it *would* land heads about half the time, and then use that counterfactual but justified claim to justify the assertion that the probability of heads on the unique toss of the specified coin is a half.

In general there are many ways of tying an event to a sequence of events whose stochastic properties are directly or indirectly known. The problem of fixing on a single way is the problem of the reference class [Kyburg, 1983]. We cannot pretend that this problem has been definitively solved, but it is quite clear that until it is, it is premature to claim that there is a lack of statistics relevant to any given sentence.

### 3.2 Inconsistency.

A more interesting — and also more problematic — issue concerns consistency. The "lottery paradox" [Kyburg, 1961] shows that however demanding we make the threshold for probabilistic acceptance, inconsistency threatens. The story runs as follows. Choose any high degree of probability — say $1-\epsilon$ — as a sufficiently high degree of probability for acceptance (for the class of contexts with which we are concerned.) Now imagine a lottery with $\lceil 1/\epsilon \rceil$ tickets. Put what conditions you will on the lottery to ensure that it is fair, and suppose that it is reasonable for us to accept those conditions. Then the probability that a specified ticket (say ticket #139076) will lose is $1 - [\lceil 1/\epsilon \rceil]^{-1}$. But this is at least as large as $1-\epsilon$, and so we should be entitled to accept, on grounds of high probability, the proposition that ticket #139076 will lose. Exactly the same argument will hold for any other ticket. Combined with the most obvious fairness constraint, that at least one ticket will win, these $\lceil 1/\epsilon \rceil$ statements are inconsistent.

A number of responses to this oddity have been proposed. Most of them have taken the form of holding to the demand that the set of statements we accept be consistent, and adding conditions to the probabilistic acceptance rule in order to ensure that this demand is satisfied.

Keith Lehrer [Lehrer, 1975] proposed that we ensure consistency by allowing the acceptance of a high probability sentence $S$ only when its probability is positively higher than that of any alternative. Thus a probability of $1 - \epsilon$ is sufficient for acceptance only if it is higher than the probability of any sentence contrary to $S$. This has some odd consequences. Consider a biased lottery, in which each of the $N$ tickets has a slightly different probability of being the winner. Without loss of generality, suppose that the probability of the $i^{th}$ ticket is less than that of the $i + 1^{st}$. Then we can be sure that the first ticket will lose. Accepting that the first ticket will lose, we can be sure that the second ticket will lose. Accepting that the second ticket will lose, ...., we can finally be sure that the every ticket but the $N^{th}$ ticket will lose, and thus that the $N^{th}$ ticket will win. We have preserved consistency, but only with a loss of generality (we can no longer deal with the equiprobable case) and at a cost of implausibility: the probability that the $N^{th}$ ticket will win may be extremely low; yet we may accept it!

John Pollock [Pollock, 1990] offers a different solution. Like Lehrer, he thinks we should not accept any of the statements of the form ticket $i$ will lose in the fair lottery. But he locates the trouble in stochastic dependence. That ticket $i$ loses *increases* the probability that ticket $j$ will win. Pollock therefore draws a distinction between the paradox of the preface [Makinson, 1965] and the lottery paradox. But as Goodwin and Neufeld have shown [Goodwin and Neufeld, 1996], many State and Provincial lotteries have a structure that supports a lottery paradox argument despite independence of the tickets. The distinction between the lottery and the preface doesn't do what Pollock wants it to do.

Isaac Levi [Levi, 1967] adopts principles that assure that both high probability and consistency are assured. But the requirement of consistency is built quite directly into his acceptance rules.

One simple possibility is to accept statements whose probabilities are high, so long as they do not introduce inconsistency. This makes the set of statements that are accepted depend on the order in which statements are considered. If we start with ticket #1, then we can accept the claim that it will lose; but if ticket #1 is considered last, then we cannot accept that claim. Note that in the case of the classical lottery, the set of statements accepted, for a given ordering of the lottery tickets, will constitute a complete description of the outcome of the lottery: of each of the tickets but one, we will accept that the ticket loses; and of the last ticket, in virtue of the fact that we can accept that at least one ticket wins, we will be sure that it wins.

Teng [Teng, 1996b] provides a treatment that avoids this problem by taking account of the accepted statements in computing the probability of a given statement. Thus we accept the statement that ticket $i$ will not win. Then we accept the statement that ticket $j$ will not win only if the probability that ticket $j$ will lose, *given* that ticket $i$ loses, is over the threshold. This amounts to adopting the same fixed point idea that inspires default logic: We accept what is probable relative to what we have accepted. This procedure has a nice semantic characterization in terms of Teng models [Teng, 1996a].

In a number of these treatments, particularly in the last two, we see a problem that is called in default logic the problem of multiple extensions. What you can accept, what you can believe, depends on the order in which you consider the candidates for belief. If all extensions are to be taken conjunctively, then of course we are back in the world of inconsistency. If they are to be taken disjunctively, then, at least in the example of the lottery, we are back in the world of evidence: we have not allowed ourselves to make any nonmonotonic inferences at all!

## 4 Embracing the Absurd.

There are a number of formalisms [Priest *et al.*, 1989a; Priest, 1989; da Costa, 1974; da Costa *et al.*, 1990; Priest *et al.*, 1989b; Rescher and Brandom, 1979; Schotch and Jennings, 1989] in which to accept a set of inconsistent premises is not a total disaster. Many of these formalisms are focused on more difficult and deeper problems than face us in making sense of probabilistic acceptance.

### 4.1 Strong and Weak Inconsistency.

There are two senses that may be given to inconsistency. In the strong sense, my beliefs, the set of propositions that I fully accept, are inconsistent when there is a self-contradictory statement among them: a statement of the form $S \wedge \neg S$. I am guilty of this when I assert in the same breath that it is raining and that it is not raining. As this example shows, it is possible to make sense of such assertions, and some of the writers mentioned attempt to do just this. ("In a sense it is raining, but in another sense it isn't.")

For our purposes this strong form of inconsistency can be disregarded. It can (surely) never be the case that the statement $S \wedge \neg S$ is highly probable.

The sense of inconsistency that threatens to follow from probabilistic acceptance is much weaker than this. Inconsistency in this weak sense characterizes a set of statements that entails a contradiction, a set of statements that admits of no model. Probabilistic acceptance, it is clear, only leads to inconsistent beliefs in this weak sense. Furthermore, reflection suggests that this is not altogether surprising. Consider any large database. It is easy enough to ensure that no explicit contradiction is directly entered into it; it is quite another matter to ensure that the entire set of entries is perfectly consistent. Consider your own personal body of knowledge. Are you quite sure that within it lurks no set of statements that is jointly unsatisfiable? Or, finally, consider scientific measurement. We can be sure of each measurement that it is accurate within three standard deviations. We can be equally sure that a thousand or so will contain at least one measurement that is not accurate within three standard deviations.

What can we do about such inconsistencies? The standard response is to root them out. When you find an inconsistency in a database, you attempt to find the guilty entry; when you find yourself caught in an inconsistency, you try to find its source and expunge it; ... but this doesn't seem to apply to the interesting and valuable case of measurement: often you can't settle on any *particular* measurement to reject.

Let us focus on inconsistent bodies of statements in somewhat more detail.

### 4.2  Finer Distinctions.

First let us note that absent strong inconsistency, we face no difficulties in adding to an inconsistent set of statement logical consequences of each particular member of that set. That is, if $S$ is a member of the set of accepted statements, and $S$ entails $T$, then we are no worse off than we were before if we add $T$ to the set of statements. Furthermore, if the set of statements is the result of probabilistic acceptance, $T$ is already there:

**Theorem 1** *If $S$ is the set of statements whose probability is greater than $1 - \epsilon$, $T \in S$, and $T \vdash W$, then $W \in S$.*

We are thus perfectly free to apply logic to single statements in $S$; we won't get anything that is not also justified probabilistically. How about using more than one premise? Suppose our level of acceptance is $1 - \epsilon$. Then any statement entailed by $k$ premises will also be entailed by the conjunction of the $k$ premises, and must therefore be as probable as that conjunction. But the probability of the conjunction of $k$ statements whose probabilities exceed $1 - \epsilon$ is at least $1 - k\epsilon$.

**Theorem 2** *If $P(A_i) \geq 1 - \epsilon$ for $1 \leq i \leq k$, then $P(A_1 \wedge A_2 \wedge \ldots \wedge A_k) \geq 1 - k\epsilon$.*

Proof. $P(A_1 \wedge A_2 \wedge \ldots \wedge A_k) = 1 - P(\overline{A_1} \vee \overline{A_2} \vee \ldots \vee \overline{A_k})$. But $P(\overline{A_1} \vee \overline{A_2} \vee \ldots \vee \overline{A_k}) \leq \sum P(\overline{A_i})$, so that $P(A_1 \wedge A_2 \wedge \ldots \wedge A_k) \geq 1 - k\epsilon$. (Note that we are making no assumptions about independence.) ∎

It is a corollary of this theorem that it takes $\lceil 1/\epsilon \rceil$ premises to derive a contradiction (or even a sentence whose probability is 0!) from a set of sentences accepted on the basis of high probability $1 - \epsilon$.

There is thus a close connection between conjunctive closure and deductive closure. We can accept at the level $1 - k\epsilon$ the conjunction of any $k$ sentences we accept at the level $1 - \epsilon$. We can accept at the level $1 - k\epsilon$ the deductive consequences of any $k$ premises we can accept at the $1 - \epsilon$ level.

### 4.3  Deductive Closure

We may take advantage of a limited amount of deductive closure within sets of statements accepted on the basis of high probability, even though these sets are weakly inconsistent. We may accept the logical consequences of any single statement that is accepted; by reducing the acceptance level to $1 - k\epsilon$ we may take account of the consequences of $k$ acceptable premises.

How has inference been circumscribed in the world of paraconsistent logic, where inconsistent sets of premises are also taken seriously? One view considers the closures of maximal consistent subsets of the inconsistent set $\mathcal{S}$[Schotch and Jennings, 1989]. The "degree of inconsistency" of a set of statements is indicated by the number of maximal consistent subsets it takes to capture all the statements of $\mathcal{S}$. In the case of an $n$-ticket lottery, for example, there are $n$ maximal consistent subsets, each corresponding to a detailed scenario of the outcome of the lottery. The logical closure of a maximal consistent subsets of $\mathcal{S}$ corresponds quite straight-forwardly to an *extensions* of nonmonotonic logic.

Of course, while these logical closures are consistent, they are also pretty far fetched: they correspond to improbable stories. The same may be true of the extensions of nonmonotonic logic. This doesn't make them totally useless: they represent the way things *could* be, and that may be worth taking account of. It is not the same, however, as taking something, tentatively, to be the case, in the sense that you could act on it.

In [Kyburg, 1974] bodies of knowledge were allowed to be inconsistent and deductively closed maximal consistent subsets of these bodies of knowledge were used as an auxiliary construction for defining randomness and, subsequently, probability. But there was no suggestion that *strands*, as these constructions were called, served any other important epistemic purpose. They were not, for example, to be construed as comprising a set of practical certainties.

## 5    Classical Statistics and Acceptance.

Classical statistical inference takes as its fundamental mechanism the rejection of a statistical hypothesis under certain prespecified conditions. Although in classical statistics probability is emphatically identified with frequencies [Neyman, 1950], or the conceptual counterparts of frequencies [Cramer, 1951], or set measures in a sample space [Lindgren, 1976], and although many statisticians strongly deny that "rejecting" $H$ amounts to "accepting" $\neg H$, much of classical statistics translates smoothly into the formalism of purely probabilistic acceptance. Furthermore, since statistical knowledge is both uncertain, and central to the application of classical decision theory, this represents an important tie between the use of probabilities in decision, and the importance of nonmonotonic inference in setting the parameters within which decision theory operates.

The core of classical statistical testing is this [Fisher, 1956]: Suppose that $H$ is a statistical hypothesis, and that $F$ is a set of possible results of observations. Under suitable circumstances we can find a region $R$ in $S$ such that *if* the hypothesis is true, then we will falsely reject $H$ only rarely by adopting the rule that we should reject $H$ just in case we make an observation falling in $R$. The general idea is that the test has the long run property that if it is applied it will lead to false rejection with a frequency less than $\epsilon$. There are more complicated situations that can be considered, (choosing between two classes of hypotheses, for example) and more complicated tests that can be analyzed (for example mixed tests in which the rejection of a hypothesis depends not only on the evidence, but also on the outcome of a "chance event") but the essence of the classical view can be captured by a simple test.

When the statistician has found a test with "nice" long run properties, he is done. The next step is a practical one: we draw the sample, obtain an element $s$ of $F$, and discover that in fact it is in the rejection region $R$ of $F$. It is at this point that we go beyond what is classically permitted. We are permitted to say we obtained a sample falling into $R$, and that $R$ has such and such nice properties. We are *not* permitted to say that, relative to this evidence, the probability that $H$ is false is less than $\epsilon$. As Birnbaum [Birnbaum, 1969] has pointed out, it is easy to avoid saying this, but it is very hard not to think it.

There are some cautions, however. There is a gap between a long run frequency and epistemic probability. An event (taking a sample and having it fall in $R$) can fall into many classes about which we have some frequentistic knowledge. We must choose the right such class, and our epistemic probability must be determined by what we know about that class. There are cases — for example such cases arise in epidemiology — where determining this class can be controversial. What is involved here is the problem of choosing the right reference class, a knotty and unsolved problem that has only begun to be explored [Kyburg, 1983].

Furthermore we must take account of the niceties of negation. While the statistician is perfectly correct in pointing out that to *fail* to *reject* a hypothesis is not at all the same as to accept the hypothesis, rejecting a hypothesis and accepting its negation may amount to different things because we may be thinking, for example, of a particular alternative to that hypothesis rather than its bare logical negation.

Leaving to one side these niceties, and speaking as ordinary scientists, we *do* accept the negations of hypotheses rejected at the .01 level — corresponding to the outcome of tests that will lead to false rejection no more than 1% of the time.

Let us now reflect on a sequence of such tests. To fix our ideas, suppose there are two, leading me to reject $H$ at the .01 level, and to reject $K$ at the .01 level. If rejecting $H$ is accepting $\neg H$ and rejecting $K$ is accepting $\neg K$, then what should our epistemic stance toward $\neg H \wedge \neg K$ be? Surely we should not be quite so confident in rejecting both $H$ and $K$ as we are in rejecting each of them separately.

Our previous considerations suggest that "full belief" in the conjunction should be characterized by $1 - .02$ rather than $1 - .01$. The ratio of costs to benefits that result from acting on the assumption of $\neg H$ may range from 1:100 to 100:1; the ratio of costs to benefits that result from acting on the assumption of $\neg H \wedge \neg K$ is restricted to the range 1:50 to 50:1. Any deductive consequence of the conjunction of $\neg H$ and $\neg K$ should also be taken to be supported to a smaller degree than either $\neg H$ or $\neg K$. If the two hypotheses are *independent* then the support for the conjunction of the negations is $1 - .02 + .0001 = (1 - .01)^2$; note that we do not gain very much from independence.

This brief discussion is not intended to do more than hint at connections between nonmonotonic inference and classical statistical inference. The point is that some of the same issues arise.

Bayesian statistical inference is different. As we already pointed out, in principle the Bayesian can incorporate the uncertainty distributions associated with the relevant statistical hypotheses into the probabilistic considerations peculiar to a specific decision context. There is, however, a computational cost involved. For present purposes, we have chosen to stand aside from the Bayesian/non-Bayesian debate. (See [Cheeseman, 1988] and [Kyburg, 1994].)

## 6  Problems and Questions

We are left with a number of important questions.

1. Assuming that acceptance is useful on some occasions, it remains to be seen whether purely probabilistic acceptance, as outlined here, is as useful as other forms of nonmonotonic acceptance. At the very least, it seems to call attention to issues that should be faced by any nonmonotonic logic: especially, the issue of what epistemic stance we should adopt toward distinct extensions of the same nonmonotonic base.

2. Assuming that there are occasions where rules leading to acceptance are useful, can we characterize those occasions succinctly? Are there situations that call for a specific nonmonotonic mechanism and others that call for other mechanisms? In view of the interreduction of various approaches to nonmonotonic inference, it seems intuitively unlikely that there are important differences where one logic rather than another is called for.

3. If there are occasions where purely probabilistic rules of acceptance serve an important function, can we find ways of characterizing these occasions?

4. What is the relation between probabilistic acceptance rules and the decision-theoretic approach. Are there situations in which they come into conflict?

5. Can we exhibit real (and not merely realistic) cases in which combining acceptance and probability leads to significant savings in computational cost or a serious improvement in performance? It would be interesting to look at cases complicated enough that we can't "see" the answer, but simple enough that we could run comparisons between a system that makes use of tentative acceptance, and a purely probabilistic system, if that is possible.

6. Statistical knowledge is clearly central to any form of decision making. This is just the sort of knowledge that we should be able to incorporate nonmonotonically into our bodies of knowledge. Thus our nonmonotonic handbook had better include some chapters on statistical inference. This in turn requires our giving statistics a closer examination, and an examination of a different character, than has been our wont in AI.

7. There have to be some rules about probability functions. Common sense does NOT endorse arbitrary probability functions. The relation among statistical inference, evidence, and probability distributions is complex and needs investigation.

## References

[Adams, 1986] Ernest W. Adams. On the logic of high probability. *Journal of Philosophical Logic*, 15:255–279., 1986.

[Birnbaum, 1969] Alan Birnbaum. Concepts of statistical evidence. In Morgenbesser et al, editor, *Philosophy Science and Method,*, pages 112–143. St. Martin's Press, New York, 1969.

[Braithwaite, 1946] Richard B. Braithwaite. Belief and action. *Proceedings of the Aristotelian Society*, 20:1–19, 1946.

[Cheeseman, 1988] Peter Cheeseman. Inquiry into computer understanding. *Computational Intelligence*, 4:58–66., 1988.

[Cramer, 1951] Harald Cramer. *Mathematical Methods of Statistics*, volume Princeton. Princeton Uniersity Press, 1951.

[da Costa et al., 1990] Newton C. A. da Costa, James J. Henschen, Lawrence J.and Lu, and V. S. Subrahmanian. Automatic theorem proving in paraconsistent logics: Theory and implementation. In M. E. Stickel, editor, *10'th International Conference on Automated Deduction*, pages 72–86, Berlin, 1990. Springer-Verlag.

[da Costa, 1974] Newton C. A. da Costa. On the theory of inconsistent formal systems. *Notre Dame Journal of Formal Logic*, 15:497–510, 1974.

[Fisher, 1956] Ronald A. Fisher. *Statistical Methods and Scientific Inference*. Hafner Publishing Co., New York, 1956.

[Geffner and Pearl, 1990] Hector Geffner and Judea Pearl. A framework for reasoning with defaults. In H. Kyburg, R. Loui, and G. Carlson, editors, *Knowledge Representation and Defeasible Reasoning*, pages 69–87. Kluwer, 1990.

[Goodwin and Neufeld, 1996] Scott Goodwin and Eric Neufeld. The 6-49 lottery. To appear., 1996.

[Kyburg, 1961] Henry E. Kyburg, Jr. *Probability and the Logic of Rational Belief.* Wesleyan University Press, Middletown, 1961.

[Kyburg, 1974] Henry E. Kyburg, Jr. *The Logical Foundations of Statistical Inference Reidel.* Reidel, Dordrecht, 1974.

[Kyburg, 1983] Henry E. Kyburg, Jr. The reference class. *Philosophy of Science*, 50:374-397., 1983.

[Kyburg, 1988] Henry E. Kyburg, Jr. Full belief. *Theory and Decision*, 25:137-162., 1988.

[Kyburg, 1994] Henry E. Kyburg, Jr. Believing on the basis of evidence. *Computational Intelligence*, 10:3-20, 1994.

[Lehrer, 1975] Keith Lehrer. Induction, rational acceptance, and minimally inconsistent sets. In Maxwell and Anderson, editors, *Minnesota Studies in the Philosophy of Science VI*, pages 295-323. University of Minnesota Press, Minneapolis, 1975.

[Levi and Morgenbesser, 1964] Isaac Levi and Sidney Morgenbesser. Belief and disposition. *American Philosophical Quarterly*, 1:221-232, 1964.

[Levi, 1967] Isaac Levi. *Gambling with Truth.* Knopf, New York., 1967.

[Lindgren, 1976] Bernard W. Lindgren. *Statistical Theory.* MacMillan, New York, 1976.

[Loui, 100 106] Ronald P. Loui. Defeat among arguments: A system of defeasible inference. *Computational Intelligence*, 1988, 100-106.

[Makinson, 1965] David C. Makinson. The paradox of the preface. *Analysis*, 25:205-207, 1965.

[McCarthy and Hayes, 1969] John McCarthy and Pat Hayes. Some philosphical problems from the standpoint of artificial inteligence. *Machine Intelligence*, 4:463-502, 1969.

[McCarthy, 1980] John McCarthy. Circumscription – a form of non-monotonic reasoning. *Artificial Intelligence*, 13:27-39, 1980.

[Moore, 1985] Robert C. Moore. Semantical considerations on nonmonotonic logic. *Artificial Intelligence*, 25:75-94, 1985.

[Neyman, 1950] Jerzy Neyman. *First Course In Probability And Statistics*, volume New York,. Henry Holt And Co, 1950.

[Pearl, 1992] Judea Pearl. Probabilistic semantics for nonmonotonic reasoning. In Robert Cummins and John Pollock, editors, *Philosophy and AI.* MIT Press, 1992.

[Pollock, 1987] John Pollock. Defeasible reasoning. *Cognitive Science*, 11:481-518, 1987.

[Pollock, 1990] John L. Pollock. *Nomic Probability and the Foundations of Induction.* Oxford University Press, New York, 1990.

[Poole, 1991] David Poole. The effect of knowledge on belief: Conditioning, specificity and thelottery paradox in default reasoning. *Artificial Intelligence*, 49:281-307, 1991.

[Priest et al., 1989a] Graham Priest, R. Routley, and J. Norman. *Paraconsistent Logic: Essays on the Inconsistent.* Philosophia Verlag, Hamden, Munchen, Wien, 1989.

[Priest et al., 1989b] Graham Priest, R. Routley, and J. Norman. *Paraconsistent Logic: Essays on the Inconsistent.* Philosophia Verlag, Hamden, Munchen, Wien, 1989.

[Priest, 1989] Graham Priest. Reasoning about truth. *Artificial Intelligence*, 39:231-244, 1989.

[Reiter, 1980] R. Reiter. A logic for default reasoning. *Artificial Intelligence*, 13:81-132, 1980.

[Rescher and Brandom, 1979] Nicholas Rescher and Robert Brandom. *The Logic of Inconsistency.* Rowman and Littlefield, Totowa, 1979.

[Schotch and Jennings, 1989] Peter K. Schotch and R. E. Jennings. *On Detonating*, pages 306-327. Philosophia Verlag, Hamden, Munchen, Wien, 1989.

[Teng, 1996a] Choh Man Teng. Combining default logic and autoepistemic logic. In Grigoris Antoniou, editor, *Workshop on Nonmonotonic Logic*, pages 133-147, 1996.

[Teng, 1996b] Choh Man Teng. Possible world partition sequences: A unifying framework for uncertain reasoning. In Eric Hlorvitz and Finn Jensen, editors, *Uncertainty in Artificial Intelligence*, pages 517-524, San Francisco, 1996. Morgan Kaufman.